\documentclass[]{article}
\usepackage[a4paper]{geometry}
\usepackage{mtsummit2023}
\usepackage{times}
\usepackage{url}
\usepackage{latexsym}
\usepackage{natbib}
\usepackage{layout}
\usepackage[hang,marginal]{footmisc}
\usepackage{multirow}

\usepackage{times}
\usepackage{latexsym}

\usepackage{microtype}

\usepackage{multirow}
\usepackage{booktabs}
\usepackage[normalem]{ulem} 
\usepackage[utf8]{inputenc}
\usepackage[T1]{fontenc}
\usepackage[noabbrev,capitalize]{cleveref}
\usepackage[disable]{todonotes}

\usepackage{enumitem}
\def\XXX#1{{\textcolor{red}{XXX #1}}}

\usepackage{booktabs, multirow} 
\usepackage{soul}
\usepackage{changepage,threeparttable} 
 \definecolor{darkgreen}{rgb}{0.0, 0.7, 0.1}

\usepackage{times}
\usepackage{latexsym}

\usepackage[T1]{fontenc}

\usepackage[utf8]{inputenc}

\usepackage{microtype}

\usepackage{inconsolata}

\begin{document}

\title{\bf Negative Lexical Constraints \\in Neural Machine Translation}

\author{\name{\bf Josef Jon } \hfill  \addr{jon@ufal.mff.cuni.cz}\\
\name{\bf  Dušan Variš} \hfill  \addr{varis@ufal.mff.cuni.cz} \\ 
\name{\bf  Michal Novák} \hfill  \addr{mnovak@ufal.mff.cuni.cz} \\ 
\name{\bf  João Paulo Aires } \hfill  \addr{aires@ufal.mff.cuni.cz} \\ 
\name{\bf  Ondřej Bojar} \hfill  \addr{bojar@ufal.mff.cuni.cz} \\ 
\addr{Charles University, Faculty of Mathematics and Physics, Institute of Formal and Applied Linguistics, Prague, Czech Republic}
}

\maketitle
\pagestyle{empty}

\begin{abstract}
This paper explores negative lexical  constraining in English to Czech neural machine translation. Negative lexical constraining is used to prohibit certain words or expressions in the translation produced by the neural translation model. We compared various methods based on modifying either  the decoding process or the training data. The comparison was performed on two tasks: paraphrasing and feedback-based translation refinement.  We also studied to which extent these methods ``evade" the constraints presented to the model (usually in the dictionary form) by generating a different surface form of a given constraint.
We propose a way to mitigate the issue through training with stemmed negative constraints to counter the model's ability to induce a variety of the surface forms of a word that can result in bypassing the constraint. We demonstrate that our method improves the constraining, although the problem still persists in many cases.
\end{abstract}

\section{Introduction}
\label{sec:introduction}
In general, lexically constrained neural machine translation (NMT) is a method that allows enforcing presence or absence of certain words or phrases in the translation output \todo{previous work ref. with more details on the task?}.
Positively constrained translation is more common and is used, for example, in named entities translation \citep{li-neural-2019,yan-impact-2019}, terminology integration \citep{dinu-etal-2019-training, jon-etal-2021-end}, or interactive machine translation \citep{knowles-koehn-2016-neural}.

Negative constraining serves different purposes.
In this paper, we focus on two use-cases: (1) paraphrase generation and (2) refining translation based on feedback.
Paraphrasing aims to produce a new translation hypothesis that differs from the original translation without significant changes in meaning. 
On the other hand,  translation refinement involves replacing specific tokens in the original translation.
These tokens can be selected either manually by the user or automatically using techniques like word-level quality estimation \citep{kepler-etal-2019-openkiwi}.
\todo{citation of QE?}
Negative constraining is particularly well-suited for translation refinement, while it can be one of the solutions for paraphrase generation.

After providing a summary of related work (\cref{sec:const_translation}), we proceed to describe the two tasks in detail (\cref{task-description}).
Next, we delve into the methods we employ to achieve negative constraining (\cref{sec:methods}). 
The results are presented in \cref{sec:exp}, followed by a manual analysis of the outputs in \cref{sec:analysis}.

\section{Related work}
\label{sec:const_translation}
There are three dominant approaches to constrained NMT. The earliest ones were based on replacing the constrained expressions in the source sentence with placeholders, ensuring that the placeholders are copied into the translation produced by the model and, finally, replacing the placeholders in the target with the desired expression \citep{crego2016systrans,hanneman-dinu:2020:WMT}.

The second class of methods is based on modifying the decoding mechanism in such way that only translations including (or not including) the specified words or phrases can be produced in the final output \citep{anderson-etal-2017-guided, hasler-etal-2018-neural,chatterjee-etal-2017-guiding, hokamp-liu-2017-lexically, post-vilar-2018-fast,hu-etal-2019-improved}.

The third class of methods revolves around altering the source input in the training data, allowing the NMT model to learn how to incorporate the constraints.
This is typically done by either appending the constraints to the end of the source sentence as a suffix or intertwining them with the source sentence and distinguishing them from its tokens using factors \citep{dinu-etal-2019-training,song-etal-2019-code, chen-etal-2020-lexical,jon-etal-2021-end,bergmanis-pinnis-2021-facilitating,bergmanis-pinnis-2021-dynamic}.

Currently, most of the research in the field focuses on positive lexical constraints, often used for terminology integration.
In contrast, there is a relatively less emphasis on negative constraining, despite its applications in areas like paraphrase generation \citep{hu2019parabank, kajiwara-2019-negative}.
These works apply a method developed by \citet{post-vilar-2018-fast} and later improved by \citet{hu-etal-2019-improved}.
This method modifies the beam search decoding algorithm so that the beam in each time step includes the best hypotheses that satisfy from zero to the full number of pre-defined constraints.
When using only negative constraints, the algorithm effectively boils down to filtering out hypotheses that would introduce any word (or phrase) from the list of constraints.

\section{Task description}
\label{task-description}

We carry out experiments with negative constraints in the two following tasks:


\paragraph{Paraphrase generation} is often achieved through translation, where negative constraints come in handy for indicating the desired differences in the paraphrased output.
To create a paraphrase of a source sentence, we go through multiple rounds of translation, each time disallowing some of the words generated in the previous pass.
These restricted words or expressions should be replaced by synonymous expressions by the MT model, thereby creating a paraphrase of the original translation.
As an example, consider the sentence \textit{``He dodged the ball.''} as the initial translation from a foreign language into English.
When the word \textit{``dodge''} is employed as the negative constraint, the system is expected to generate a paraphrase of the original translation (e.g. \textit{``He avoided the ball.''}) in the second pass.

\paragraph{Feedback-based translation refinement} involves using external feedback to assess the model's output, for example, through user feedback in an interactive setting.
After the initial translation is presented, the user can identify certain words as mistranslated.
%
These words are then excluded from the subsequent output, prompting the model to generate a potentially improved translation.
As obtaining human constraints can be costly, we translate the source without any constraints and analyze the tokens present in the MT output but not in the reference. 
In the next translation pass, we constrain the model to avoid using these ``unconfirmed'' tokens and evaluate the resulting translation.

In practice, word-level quality estimation (QE) systems can partially replace user feedback by highlighting potentially problematic tokens.
In our work, we use references as a proxy for an oracle QE.

\section{Proposed methods}
\label{sec:methods}

We define a constraint as a sequence of consecutive subwords, which may represent either a single word or a multi-word expression.
Each input example can have a list of multiple constraints that need to be satisfied.
To incorporate these constraints into the translation process, we implement the following methods.

\paragraph{Beam filtering}
This method is based on an existing implementation where a hypothesis containing any forbidden subword is dropped from the beam search.\footnote{Implemented here: \url{https://github.com/XapaJIaMnu/marian-dev/tree/paraphrases_v2}}
For each input sentence, a list of constraints (where each constraint represents a single subword) is provided.
During beam search, any time a hypothesis that contain a constraint from the list is generated, it is removed.
Optionally, it is removed only if the log probability of the subword is falls below a specified threshold.
This method is referred to as the ``subword method'', and we extend it to support multi-subword expressions (``multi-subword method''). Instead of filtering after a single subword is generated, we store subwords corresponding to each constraint in a list of lists.
For example:

{\footnotesize
\begin{itemize}[noitemsep]
    \item \textbf{Constraint 1:} decoding  \textbf{Segmentation: }\_deco ding
    \item \textbf{Constraint 2:} beam search \textbf{Segmentation:} \_be am \_search
    \item \textbf{Subword method:} [\_deco, ding, \_be, am, \_search]
    \item   \textbf{Multi-subword method:} [[\_deco, ding], [\_be, am, \_search]]
 
\end{itemize}
}

Each hypothesis tracks its progress through the constraints, and it is removed only when a complete constraint is met.
In other words, the hypothesis is removed only when all the subwords forming a single constraint are generated subsequently.\footnote{Link to the github repository of our code, removed for review. 
}

\paragraph{Score penalty}
Another technique we experimented with is modifying the output probability of the subwords that form the constrained expression during the decoding.
For this technique, we provided a list of constraints along with each input sentence.
We created a mask with a penalty value for each subword present in the vocabulary.
In our implementation, the penalty value was global, meaning each subword had either no or the same specified penalty.
This mask was then summed with the output logits at each decoding step.
To handle multi-subword constraints, we used a trie structure to track the progress through each constraint in each beam, similar to the approach used in  \citep{hu-etal-2019-improved}.


In the trie structure, each node represents a subword that is part of a constraint.
The node contains a list of vocabulary IDs that, if generated in the next decoding step, would complete the constraint.
When the subword represented by a node is produced, the penalty is added to the scores of these IDs in the next step.


\paragraph{Learned constraints}
A different approach to constraining involves modifying the training data to bias the model.
The objective is to prevent the model from producing the constraint expressions that are directly provided with the input sentence.
In our experiments, we separate the list of constraints from the source sentence by a special \texttt{<sep>} token, whereas the individual constraints within the list are separated by a special \texttt{<c>} token. 
For example: 

{\footnotesize
\begin{itemize}[noitemsep]
    \item This is a sentence where we want to use synonyms for dog and cat. \texttt{<sep>} dog \texttt{<c>} cat
 
\end{itemize}
}

We train a model on the original dataset and the use this model to translate the source side of the dataset.
Tokens present in the translation but not in the reference are extracted and used as ``synthetic'' constraints for training data, similar to the approach in the \textit{Translation refinement} task. 
The resulting training dataset with ``synthetic'' constraints is then utilized to train a model capable of handling negative constraints in its input.

\section{Experiments} \label{sec:exp}

In this section, we compare the performance of the methods on the tasks presented earlier.

\subsection{Datasets and tools}
We use CzEng 2.0~\citep{kocmi-2020-announing} dataset, all the authentic parallel sentences (61M), as the training dataset.
We use WMT \texttt{newstest-2019} \citep{barrault-etal-2019-findings} and \texttt{newstest-2020} \citep{barrault-etal-2020-findings} for development and final evaluation respectively. We also used a subset of 50 examples from English-Czech \texttt{newstest-2011} which contains a large number of references (about 15M reference sentences in total, averaging 300k references per source sentence) introduced by \citet{bojar-2013-scratching} for part of the experiments.  For evaluation on this multi-reference dataset (denoted ``Multi-ref'' in the following), we randomly picked up to 1,000 references for each source sentence to compute BLEU score and 20 references to compute COMET (the COMET scores are computed separately for each reference and averaged).


We use SentencePiece~\citep{kudo-richardson-2018-sentencepiece} for subword segmentation and UDPipe~\citep{straka-strakova-2017-tokenizing} for lemmatization.
The models are trained with Marian \citep{junczys-dowmunt-etal-2018-marian} using default hyperparameters for Transformer-base architecture.
BLEU \citep{papineni-etal-2002-bleu} scores are obtained by SacreBLEU \citep{post-2018-call}.\footnote{SacreBLEU signature: BLEU+case.mixed+lang.en-cs+numrefs.1+smooth.exp+test.wmt20+tok.13a+version.1.4.14}
For COMET \citep{rei-etal-2020-comet} scores, we evaluate with the \textit{wmt20-comet-da} model.
As the references in the Multi-ref test set are tokenized, we detokenized them using Sacremoses.\footnote{\url{https://github.com/alvations/sacremoses}}

\label{sec:experiments}
\subsection{Baseline}

\begin{table}[tp]\centering
\small
\begin{tabular}{lrrrrr}\toprule
\textbf{constraints} &\multicolumn{2}{c}{\textbf{WMT20}} &\multicolumn{2}{c}{\textbf{Multi-ref}} \\\cmidrule{1-5}
&BLEU &COMET &BLEU &COMET \\\midrule
Yes &30.8 &0.6067 &46.5 &0.5971 \\
No &30.7 &0.6071 &46.7 &0.5944 \\
\bottomrule
\end{tabular}
\caption{Comparison of the baseline models trained with and without constraints present in the training data. No constraints were present in the test set, showing that even the model exposed to the input constraints can be used in a ``default'' mode (no input constraints).}
\label{tab:baseline}

\end{table}

Our baseline model is a Transformer-base trained on CzEng 2.0 with negative constraints.
This model is specifically trained to use negative constraints provided as part of the input, as described earlier in the \textit{Learned constraints} section of \cref{sec:methods}.
This approach enables more accurate comparison with other methods of incorporating constraints.
\cref{tab:baseline} illustrates that when no constraints are provided at test time, the translation quality in terms of automated metrics is similar to a vanilla model without constraints. 

\subsection{Paraphrasing}

\begin{figure}[t]
    \centering
    \begin{minipage}[b]{\linewidth}
        \centering
         \hspace*{-0.3cm}
        \includegraphics[width=0.5\linewidth]{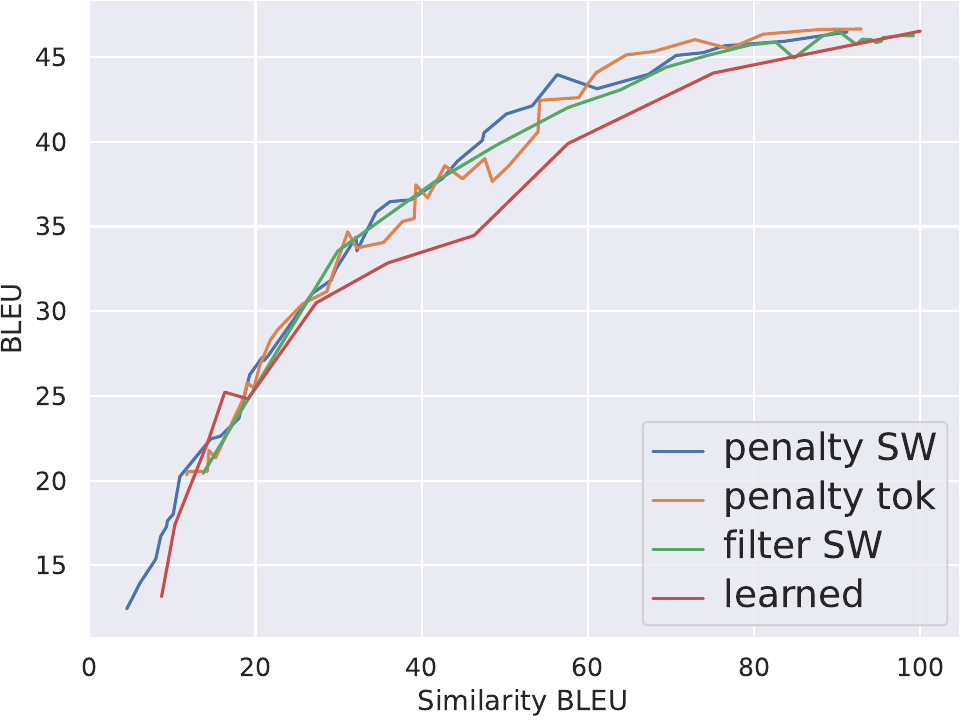}
        \includegraphics[width=0.5\linewidth]{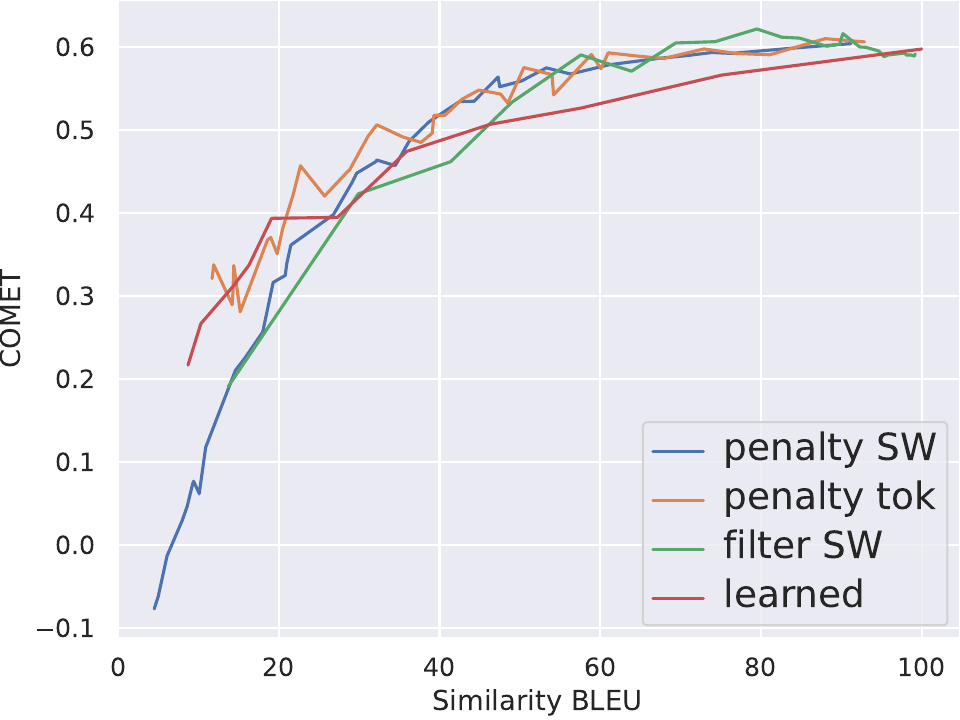}
    \end{minipage} 
    \caption{Correlation between either BLEU (left) or COMET (right) scores and similarity of translation to the baseline translation for paraphrasing.
    }
    \label{fig:correlation}
\end{figure}

In this task, our goal is to produce paraphrases that are diverse enough from the original translation.
We thus opt for a multi-reference evaluation.

We create negative constraints by translating the source sentences of Multi-ref with the baseline model.
The translations are then tokenized, removing punctuation and common Czech stopwords\footnote{Prohibiting them by a constraint would hinder generation of grammaically fluent sentences.}.
The remaining set of tokens serve as negative constraints.

In this task, our focus is on examining the relationship between the reference-based translation quality metrics (BLEU and COMET) and the similarity of the translation with the baseline translation.
The objective is to generate sentences that are as distinct as possible while minimizing the negative impact on translation quality.
The correlation for all the methods is depicted in \cref{fig:correlation}.
Sampling across a range of thresholds (see below) generates various output variants.
We arrange them on the x-axis based on their similarity with the unconstrained translation (``Similarity BLEU'').
The y-axis then represents the automatically assessed translation quality.
The curves' concave shape confirms that there is no sudden drop in quality as we paraphrase.
However, even with the very permissive scoring against the Multi-ref references, both BLEU and COMET inevitably decline as we deviate further from the initial translations.

Tables \ref{tab:para_bonus}--\ref{tab:para_learned} present the translation scores as well as the similarity of the paraphrase to the first translation (Similarity BLEU, denoted ``Sim'' here) for several thresholds.
Each threshold controls the number of tokens to be paraphrased, affecting the similarity.
However, its exact meaning differs for each method, as explained below.
%
Coverage (``Cvg'') indicates the ratio of constraint tokens that were produced in the translation (ignoring the casing). 

\begin{table*}[t]\centering
\small
\begin{tabular}{lrrrr|rrrr}\toprule
\multicolumn{5}{c}{\textbf{Single subword}} &\multicolumn{4}{c}{\textbf{Whole token}} \\\cmidrule{1-9}
\textbf{Penalty} &\textbf{$\uparrow$BLEU} &\textbf{$\downarrow$Sim} &\textbf{$\uparrow$COMET} &\textbf{$\downarrow$Cvg} &\textbf{$\uparrow$BLEU} &\textbf{$\downarrow$Sim} &\textbf{$\uparrow$COMET} &\textbf{$\downarrow$Cvg} \\\midrule
0 &46.5 &100 &0.5991 &1.00 &46.5 &100 &0.5991 &1.00 \\
0.1 &\textbf{46.5 }&\textbf{83.6} &\textbf{0.5999} &0.84 &\textbf{46.7} &\textbf{92.9} &\textbf{0.6078} &0.94 \\
0.2 &\textbf{45.9 }&\textbf{76.4} &\textbf{0.5946} &0.76 &\textbf{46.6} &\textbf{88.0} &\textbf{0.6123} &0.89 \\
0.5 &45.1 &70.6 &0.5917 &0.70 &\textbf{46.0} &\textbf{72.9 }&\textbf{0.5991} &0.73 \\
1 &41.6 &50.2 &0.5616 &0.52 &42.6 &58.9 &0.5939 &0.62 \\
2 &32.5 &29.7 &0.4469 &0.32 &35.5 &39.1 &0.4988 &0.46 \\
3 &20.2 &10.9 &0.1203 &0.18 &26.8 &20.5 &0.3869 &0.30 \\
\bottomrule
\end{tabular}
\caption{Results of the \textit{score penalty} method on the paraphrasing task. 
We boldface variants where we deem the degradation small enough (BLEU or COMET close enough to their baseline value or even better).
}
\label{tab:para_bonus}

\end{table*}

\begin{table*}[t]\centering
\small
\centering
\begin{tabular}{lrrrr|rrrr}\toprule
\multicolumn{5}{c}{\textbf{Single subword}} &\multicolumn{4}{c}{\textbf{Whole token}} \\\cmidrule{1-9}
\textbf{Thrshld} &\textbf{$\uparrow$BLEU} &\textbf{$\downarrow$Sim} &\textbf{$\uparrow$COMET} &\textbf{$\downarrow$Cvg} &\textbf{$\uparrow$BLEU} &\textbf{$\downarrow$Sim} &\textbf{$\uparrow$COMET} &\textbf{$\downarrow$Cvg} \\\midrule
0 &7.2 &2 &-0.3388 &0.07 &8.7 &2.8 &0.0621 &0.09 \\
-0.1 &20.4 &13.7 &0.1919 &0.17 &18.1 &10.5 &0.2448 &0.14 \\
-0.2 &33.5 &29.9 &0.4285 &0.37 &33.9 &26.8 &0.4595 &0.31 \\
-0.5 &42.0 &57.6 &0.5938 &0.60 &41.7 &53.3 &0.5544 &0.52 \\
-1 &\textbf{45.9} &\textbf{82.6} &\textbf{0.6146} &0.83 &45.1 &\textbf{77.8 }&\textbf{0.6059} &0.76 \\
-1.5 &\textbf{45.7 }&\textbf{92.3} &\textbf{0.6011} &0.91 &\textbf{46.1} &\textbf{89.8 }&\textbf{0.6076} &0.87 \\
-2 &\textbf{46.2} &\textbf{95.5} &\textbf{0.5901} &0.96 &\textbf{46.2} &\textbf{93.3} &\textbf{0.5774} &0.93 \\
-3 &\textbf{46.3} &\textbf{99.2} &\textbf{0.5931} &0.99 &\textbf{46.3} &\textbf{99.1} &\textbf{0.5906} &0.99 \\
\bottomrule
\end{tabular}
\caption{Results of the \textit{beam filtering} method on the paraphrasing task. Boldfacing as in \cref{tab:para_bonus}.
}
\label{tab:para_beam_filter}
\end{table*}

\begin{table}[t]\centering
\small
\begin{tabular}{lrrrrr}\toprule
\textbf{Ratio} &\textbf{BLEU} &\textbf{Sim} &\textbf{COMET} &\textbf{Cvg} \\\cmidrule{1-5}
0 &46.5 &100 &0.5991 &1.00 \\
single  &45.4 &81.4 &0.5582 &0.83 \\
0.1 &44.1 &75.1 &0.5685 &0.76 \\
0.2 &39.9 &57.6 &0.5287 &0.63 \\
0.4 &32.8 &35.9 &0.4796 &0.43 \\
0.6 &24.8 &19.1 &0.4034 &0.25 \\
0.8 &22.3 &14.3 &0.3193 &0.18 \\
1 &13.1 &8.7 &0.2194 &0.12 \\
\bottomrule
\end{tabular}   
\caption{Results of the \textit{learned} method on the paraphrasing task. We do not boldface any row because the BLEU and COMET scores immediately degrade.
}
\label{tab:para_learned}
\end{table}

The results for the \textit{score penalty} method are presented in \cref{tab:para_bonus}.
\textit{Penalty} represents the log probability that is subtracted from the logits for constrained tokens in each decoding step.
Two variants of the method are compared.
\textit{Single subword} is the simpler variant, penalizing each subword found among the constraints.
On the other hand, in the \textit{Whole token} variant, the multi-subword implementation is used.
The penalty is applied only when a whole constraint is completed in the hypothesis (in our configuration, the whole constraint will always be a single word, due to the constraint generation algorithm).
The \textit{penalty} parameter allows us to control the resulting paraphrase similarity: the higher its value, the more disadvantaged are the constrained tokens during decoding.
We observe no significant degradation of translation up until about 88 BLEU similarity (0.89 coverage).
Even at 72.9 BLEU similarity (0.73 coverage), the degradation is minimal.
Multi-subword implementation yields better results than the single-subword implementation, allowing us to reach slightly lower coverage with comparable degradation, and it even appears to improve the baseline metric levels (BLEU of 46.7 and COMET of 0.6123 instead of the baseline 46.5 and 0.5991, respectively).

For the \textit{beam filtering} method, the results are presented in  \cref{tab:para_beam_filter}.
The controlling parameter is a threshold log probability, removing the hypotheses that use the constraint with a probability below the threshold.
Opposed to the previous method, the lower its value, the more permissive the algorithm is, keeping the hypotheses with less probable constraints in the beam search.
Again, two variants (single- and multi-subword) are implemented.
For similar paraphrases, there are no notable score differences.
However, as translations become more dissimilar, the multi-subword implementation performs better.
Overall, \textit{beam filtering} and \textit{score penalty} methods show similar performance.
An improvement in overall quality in terms of COMET is again observed when deviating somewhat from the baseline output (COMET slightly above 0.60 compared to 0.59).

Results for the \textit{learned} constraints method are displayed in \cref{tab:para_learned}.
We consider content words from the baseline translation as potential negative constraints, resulting in a full set of conceivable constraints for a sentence.
The method's control parameter is the ratio of total constraints to those actually used.
For example, with 6 available constraints for a sentence and a ratio of 0.5, we select only 3 constraints.
``Singl'' in the ratio column indicates that only one constraint was used for each sentence.
The selection is based on token-level model scores from the baseline translation, where scores of subwords comprising a token are summed.
The lowest log probability tokens are constrained first, effectively preventing the usage of words that the baseline model hesitates to produce.
We chose this sampling approach after observing large result variances when using randomly sampled constraints.
However, we acknowledge that this selection method is not optimal, as several random runs led to significantly better BLEU and COMET scores.
The learned constraints underperform compared to other approaches, likely because the decoding-based methods offer more precise control over which constraints to use (penalty or threshold).


\subsection{Translation refinement}

\begin{figure}[t]
    \centering
    \begin{minipage}[b]{\linewidth}
        \centering
         \hspace*{-0.3cm}
        \includegraphics[width=0.5\linewidth]{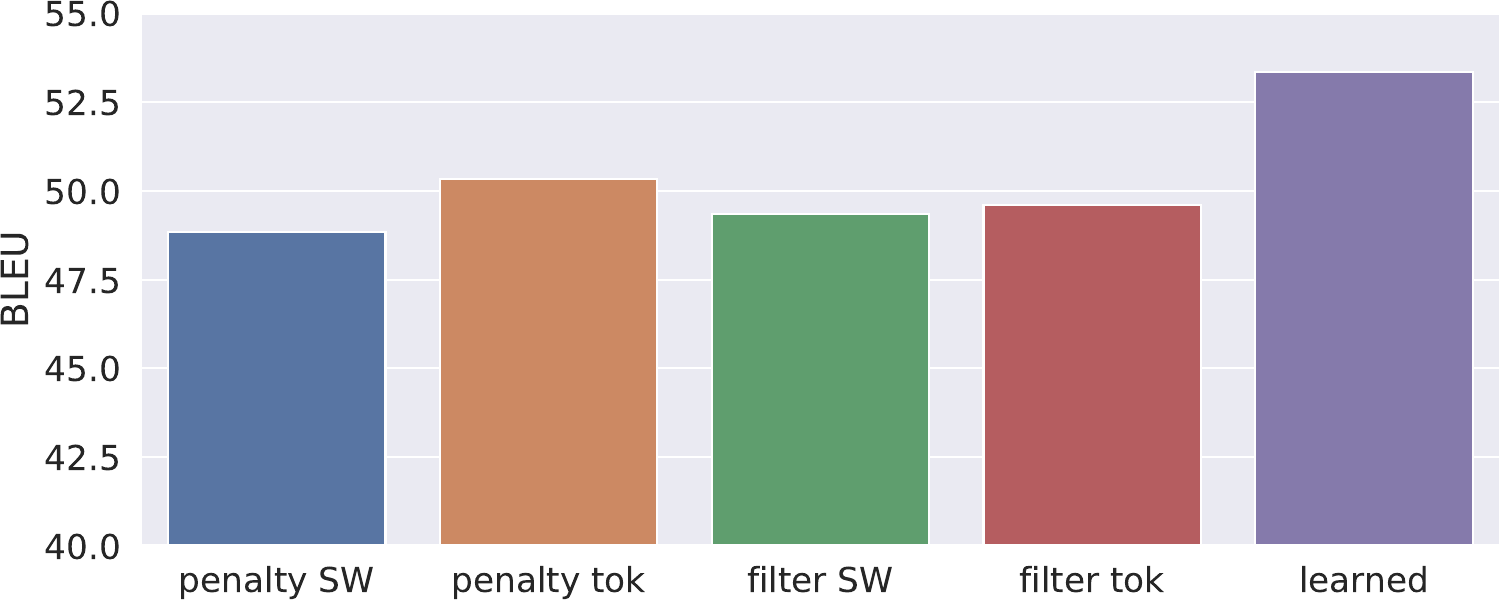}
        \includegraphics[width=0.5\linewidth]{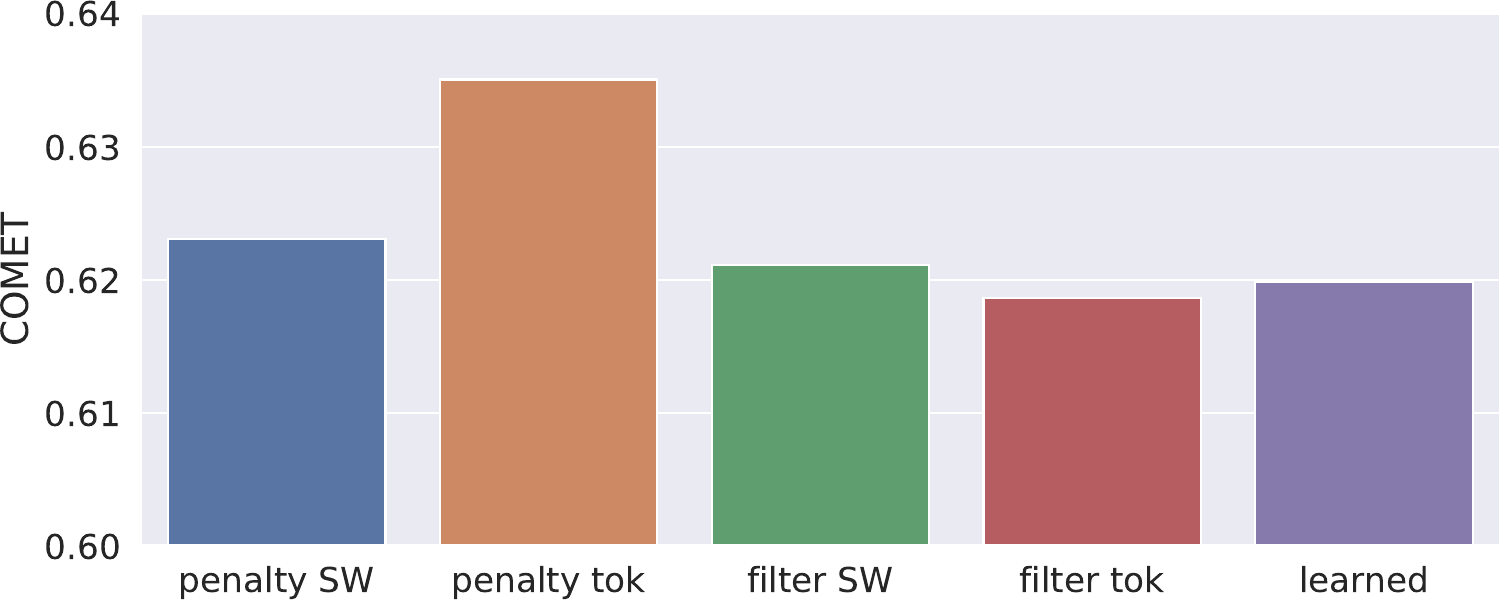}
    \end{minipage} 
    \caption{The best results obtained by each method on the \textit{translation refinement} task, either in terms of BLEU (left) or COMET (right) scores. These results were computed using the best found setting of the control parameter for each method.
    }
    \label{fig:improve}
\end{figure}

\begin{table*}[t]\centering
\small
\begin{tabular}{lrrrrrrrrr}\toprule
\multicolumn{5}{c}{\textbf{Single subword}} &\multicolumn{4}{c}{\textbf{Whole token}} \\\cmidrule{1-9}
\textbf{Penalty} &\textbf{BLEU} &\textbf{Sim} &\textbf{COMET} &\textbf{Cvg} &\textbf{BLEU} &\textbf{Sim} &\textbf{COMET} &\textbf{Cvg} \\\midrule
0 &46.5 &100 &0.5991 &1 &46.5 &100 &0.5991 &1 \\
0.1 &46.9 &95.4 &\textbf{0.6144} &0.93 &47.0 &96.5 &0.6104 &0.95 \\
0.5 &48.5 &80.6 &0.6024 &0.70 &48.7 &85.1 &0.6237 &0.76 \\
1 &\textbf{48.6} &68.9 &0.5754 &0.50 &48.5 &74.3 &\textbf{0.6302 }&0.59 \\
2 &47.1 &57 &0.5773 &0.30 &48.6 &63.9 &0.6011 &0.43 \\
3 &48.2 &53.3 &0.5617 &0.19 &49.4 &61.4 &0.5790 &0.33 \\
3.5 &48.1 &50.8 &0.5226 &0.15 &\textbf{49.4} &57.1 &0.5695 &0.22 \\
\bottomrule
\end{tabular}
\caption{Results of the \textit{score penalty} method on the refinement task.}
\label{tab:improve_bonus}
\end{table*}

\begin{table*}[t]\centering
\small
\begin{tabular}{lrrrrrrrrr}\toprule
\multicolumn{5}{c}{\textbf{Single subword}} &\multicolumn{4}{c}{\textbf{Whole token}} \\\cmidrule{1-9}
\textbf{Thrshld} &\textbf{BLEU} &\textbf{Sim} &\textbf{COMET} &\textbf{Cvg} &\textbf{BLEU} &\textbf{Sim} &\textbf{COMET} &\textbf{Cvg} \\\midrule
0 &47.4 &48.7 &0.4755 &0.03 &49.4 &50.1 &0.5771 &0.05 \\
-0.1 &47.9 &52.4 &0.6012 &0.19 &\textbf{49.6 }&53.4 &0.5814 &0.16 \\
-0.2 &48.7 &59.5 &0.6163 &0.37 &48.7 &56.7 &\textbf{0.6192} &0.31 \\
-0.3 &\textbf{49.4} &65.7 &0.5976 &0.46 &48.5 &63.7 &0.6179 &0.42 \\
-1 &47.1 &88 &\textbf{0.6100} &0.83 &47.7 &85.4 &0.6109 &0.76 \\
-2 &46.3 &96.9 &0.5932 &0.97 &46.5 &95 &0.5813 &0.93 \\
-3.5 &46.3 &99.2 &0.5931 &0.99 &46.3 &99.2 &0.5931 &0.99 \\
\bottomrule
\end{tabular}
\caption{Results of the \textit{beam filtering} method on the refinement task.}
\label{tab:improve_beam_filter}
\end{table*}

\begin{table}[t]\centering

\small
\begin{tabular}{lrrrrr}\toprule
\textbf{ratio} &\textbf{BLEU} &\textbf{Sim} &\textbf{COMET} &\textbf{Cvg} \\\cmidrule{1-5}
0 &46.5 &100 &0.5991 &1.00 \\
single &47.6 &82.3 &0.6123 &0.75 \\
0.1 &46.8 &94.4 &0.6058 &0.92 \\
0.2 &47.0 &83 &\textbf{0.6212 }&0.75 \\
0.4 &47.4 &72.5 &0.6026 &0.56 \\
0.6 &48.7 &65.7 &0.5922 &0.38 \\
0.8 &51.2 &58.8 &0.6103 &0.21 \\
1 &\textbf{53.4 }&55.4 &0.5746 &0.08 \\
\bottomrule
\end{tabular}
\caption{Results of the \textit{learned} method on the refinement task.}
\label{tab:improve_learned}
\end{table}

\begin{table}[t]
\centering
\small
\begin{tabular}{lrrrrrr}\toprule
\multicolumn{3}{c}{\textbf{Learned}} &\multicolumn{3}{c}{\textbf{Score penalty}} \\\cmidrule{1-6}
\textbf{ratio} &\textbf{BLEU} &\textbf{COMET} &\textbf{penalty} &\textbf{BLEU} &\textbf{COMET} \\\midrule
single &31.5 &\textbf{0.6183} &  0.2  & 30.6 & \textbf{0.6033} \\
1 &\textbf{38.5} &0.5973 &   0.1   &\textbf{30.8}& 0.6028 \\
\midrule
baseline &  30.9 & 0.6067 \\
\bottomrule
\end{tabular}
\caption{Results of best performing methods on newstest20. Results obtained using best-performing parameters for both metrics separately are shown.}
\label{tab:news20}
\end{table}


Unlike the paraphrasing task, where the relationship between similarity and translation quality is relevant, the translation refinement task solely aims to improve the absolute quality of translation.
The best scores achieved with optimal control parameters are presented in \cref{fig:improve}.
 
Results for \textit{score penalty} and \textit{beam filtering} methods are presented in \cref{tab:improve_bonus,tab:improve_beam_filter}, showing the similar performance to each other, as already observed in the previous task.
 
In the \textit{learned constraints} method  (\cref{tab:improve_learned}), the BLEU scores improve with an increasing ratio of constraints, while the COMET scores do not follow the same trend.

The \textit{learned constraints} method outperformed others significantly in terms of BLEU score.
The \textit{score penalty} method achieved a slightly better COMET score with the best penalty value.
We believe this is again due to the decoding methods providing more precise control over the enforcement of constraints compared to the learned method.

In \cref{tab:news20} we present results for the two best scoring methods on a better-known test set for comparison, \textit{newstest20} \citep{barrault-etal-2020-findings}.
The \textit{learned} method provides better results than the \textit{score penalty} method on this dataset.


\section{Manual analysis}
\label{sec:analysis}
\setlength{\tabcolsep}{5pt}

\begin{figure*}[t]
    \centering
    \scriptsize
    \begin{tabular}{p{0.2\linewidth}p{0.22\linewidth}p{0.125\linewidth}p{0.228\linewidth}l}
    \textbf{Source} & \textbf{Base translation} &  \textbf{Constraints}  & \textbf{Constrained translation}  & \textbf{Error} \\  \toprule

	Michael Jackson's former bodyguard has claimed the late singer cultivated some of his eccentricities with the deliberate intention of \textbf{riling up} the media. & Bývalý bodyguard Michaela Jacksona tvrdil, že zesnulý zpěvák pěstoval některé z jeho výstředností s úmyslem rozzuřit média. & bodyguard, tvrdil, pěstoval, své, výstřednosti, s, úmyslem, \textbf{rozzuřit}, média	&	Bývalý osobní strážce Michaela Jacksona tvrdí, že zesnulý zpěvák pěstuje některé z jeho výstředností se záměrem \textbf{rozzuřit} sdělovací prostředky. & Not satisfied \\
 \hline

       And Modi's government has created an uproar by instituting a national registry of citizens and setting up detention camps in the border state of Assam. & A Modiho vláda vyvolala pozdvižení zavedením národního registru občanů a zřízením zadržovacích táborů v pohraničním státě \textbf{Assam} & Modiho, vyvolala, pozdvižení, zavedením, zadržovacích, \textbf{Assam} & A Módího vláda způsobila rozruch vytvořením národního registru občanů a zřízením zajateckých táborů v pohraničním státě \textbf{Assam}. & Challenging
    
    \\
     \hline

     Neither chamber of Congress appears to have the two-thirds majority needed to override the president's opposition. & Zdá se, že ani jedna kongresová komora nemá \textbf{dvoutřetinovou} většinu potřebnou k překonání prezidentovy opozice. & kongresová, komora, \textbf{dvoutřetinovou}, překonání &  Zdá se, že ani jedna z kongresových komor nemá \textbf{dvoutřetinovou} většinu potřebnou k potlačení prezidentovy opozice.  & Reference  
\\ 
 \hline

    \_Last \_year , \_construction \_of \_Q id di y a \_" ent er tain ment \_city " \_was \_launched  \_near \_Ri y ad h. & \_Po bl í ž \_Ri já du \_byla \_v \_loňském \_roce \_zahájen a \_výstavba \textbf{\_útvar ového }\_města \_Q id di y a & \textbf{\_útvar ového} & Po bl í ž \_Ri já du \_byla \_v \_loňském \_roce \_zahájen a \_výstavba \_„ \textbf{ú t var ového }\_města “ \_Q id di y a & Segmentation \\ 
     \hline

    A Pittsburgh native whose real name was Malcolm James Myers McCormick, Miller's lyrics included frank discussion of his depression and drug use. & Domorodec z Pittsburghu, jehož pravé jméno bylo Malcolm James Myers McCormick, Millerovy texty zahrnovaly upřímnou diskusi o jeho depresi a užívání drog. &  \textbf{domorodec}, pravé, \textbf{Millerovy}, \textbf{texty}, zahrnovaly, upřímnou, \textbf{diskusi}, \textbf{depresi} & \textbf{Domorodce} z Pittsburghu, jehož skutečné jméno bylo Malcolm James Myers McCormick, \textbf{Millerův} \textbf{text} obsahoval otevřenou \textbf{diskuzi} ohledně \textbf{deprese} a užívání drog.  & Inflection 
    \end{tabular}
    \caption{Examples of baseline and constrained translations with interesting behavior. The columns show the English source sentence, baseline translation into Czech, list of constraints, and the final constrained translation. The last column contains a type of error observed.
The Segmentation example is shown in subword units for explanation purposes.
    }
    \label{tab:err_examples}
\end{figure*}

\begin{table}[t]\centering
\footnotesize
\begin{tabular}{lrrrrr}\toprule
\textbf{Model} &\textbf{Constraints} &\textbf{BLEU} &\textbf{Surface Form Cvg} &\textbf{Lemma Cvg} \\\midrule
SF &no &30.9 &1.00 &0.96 \\
SF &SF &38.5 &0.09 &0.34 \\
Stem &no &30.9 &1.00 &0.96 \\
Stem &Stem &36.9 &0.22 &0.39 \\
\bottomrule
\end{tabular}
\caption{Comparison of surface form and lemma coverage (Cvg) for models trained with either surface form or stemmed  constraints. Evaluated on \texttt{newstest-2020}.}
\label{tab:lemma}
\end{table}

Our results show that the methods tend to overlook some negative constraints and still produce prohibited words. Both the \textit{score penalty} and \textit{beam filtering} methods require pushing the thresholds quite far to satisfy all constraints. Conversely, the \textit{learned} method is more attentive to constraining but results in quick degradation of translation quality. To gain insights into the system behavior, we examined the outputs and present typical examples for each class in \cref{tab:err_examples}.
These examples are from the \textit{translation refinement} task using the \textit{learned} method, with constraints being tokens present in the baseline translation but not in the reference.
The first example showcases a clear failure of the method, as the constraint is ignored without any apparent reason.
The second example is challenging, as it requires knowledge of the Czech transcription of the name \emph{Assam} based on its English transcription.

The \textit{Reference} error example illustrates a situation, where the the meaning of the reference translation that we use to generate the negative constraints slightly deviates from the source sentence, resulting in a constraint difficult to satisfy.
The reference translation replaces the term \textit{two-thirds} (\textit{dvoutřetinovou}) with a different term, \textit{needed} (\textit{potřebnou}), which leads to \textit{dvoutřetinovou} being selected as a constraint.
Since it is difficult to translate \textit{two-thirds majority} differently from the baseline translation, the model fails to do so.
This issue could be addressed by using a validation dataset with more accurate reference translations. 

In the \textit{Segmentation} error example, the constraint is circumvented by employing a different subword segmentation of the output.
Sinve we use SentencePiece without prior tokenization, adding a quotation mark („) at the beginning of a token results in a different segmentation that is not accounted for by the constraints (as the constraints are provided to the model with pre-existing segmentation).

The \textit{Inflection} example demostrates a scenario where the model managed to avoid generating a constraint in a specific form but did not avoid producing the constrained term itself.
Out of 8 constraints, 4 are fulfilled with a different inflected form in the constrained translation (in addition, one constraint is produced with a different spelling: \textit{diskusi/diskuzi}).
This behavior is undesirable because such circumvention can still lead to a potentially problematic translation.
However, in certain cases, like paraphrasing, it may be deemed acceptable. 

The extent of this behavior is presented in \cref{tab:lemma}. 
We conduct a comparison between coverage at the surface form level and coverage at the lemma level.
The evaluation is based on the \textit{translation refinement} task on \texttt{newstest-2020}, using the \textit{learned} method with a constraint usage ratio of 1.0.
For the lemma-level coverage assessment, both the constraints and constrained translation were lemmatized.
This ensures that even when the constraint is generated in a different surface form, it is considered covered.
It is important to note that our lemmatization method is context-dependent, and in some cases, different lemmas may be produced for the same word in a sentence and in the constraint list, leading to some imprecision in these results.

At the surface level, the coverage is 0.09, indicating that 91\% of the constraints are correctly satisfied.
However, at the lemma level, the coverage increases to 0.34, which means that another 25\% of the constraints appear in the translation in a different surface form, not detected by the previous method of computing coverage.
We attempted to mitigate this behavior by training the model to use stemmed constraints (\textit{Stem} model in \cref{tab:lemma}).
Our goal was to leverage the language modeling capability of the NMT model to account for all the possible word forms.
While this approach partially works, reducing the gap between surface form and lemma coverage to 17 instead of 25, the overall performance is inferior (BLEU of 36.9 instead of 38.5).

\section{Conclusion}
\label{sec:conclusion}
We conducted a thorough investigation into NMT decoding with negative lexical constraints, addressing two tasks: paraphrasing and interactive translation refinement. Our comparison of various approaches revealed that it is indeed possible to restrict the NMT model from generating specific words in its output. However, none of the methods provided flawless results. By examining the errors made by the most effective approach, we identified instances where the model evades the constraints in morphologically rich languages by producing slightly different surface forms of the prohibited words. While we proposed a simple solution by training the model to use stemmed constraints, it adversely impacts the overall translation quality. Despite these challenges, our research sheds light on the potential of using negative constraints in NMT decoding and highlights areas for further improvement.

\section*{Acknowledgements}
This work was partially supported by the Charles University project GAUK No. 244523, the grant
825303 (Bergamot) of the European Union’s Horizon 2020 research and innovation	programme, 
the grant 19-26934X  (NEUREM3)  of  the  Czech  Science Foundation and 
the grant FW03010656 of the Technology Agency of the Czech Republic. 

\bibliography{anthology,custom}
\bibliographystyle{apalike}
\clearpage



\end{document}